\documentclass[10pt,twocolumn,letterpaper]{article}

\usepackage{cvpr}
\usepackage{times}
\usepackage{epsfig}
\usepackage{graphicx}
\usepackage{amsmath}
\usepackage{amssymb}
\usepackage{multirow}
\usepackage{siunitx}
\newcommand{\tabincell}[2]{\begin{tabular}{@{}#1@{}}#2\end{tabular}}

\usepackage{color, colortbl}
\definecolor{LightCyan}{rgb}{0.88,1,1}
\definecolor{LightRed}{rgb}{1,0.88,0.95}

\usepackage{amssymb}
\usepackage{pifont}
\newcommand{\xmark}{\ding{55}}

\usepackage[breaklinks=true,bookmarks=false]{hyperref}

\cvprfinalcopy 

\renewcommand{\thefootnote}{\fnsymbol{footnote}}

\begin{document}

\title{PV-RCNN: Point-Voxel Feature Set Abstraction for 3D Object Detection}

\author{\stepcounter{footnote}
	Shaoshuai Shi$^{1}$
	~~\quad Chaoxu Guo$^{2,3}$ 
	~~\quad Li Jiang$^4$ \\
	~~\quad Zhe Wang$^2$  
	~~\quad Jianping Shi$^2$
	~~\quad Xiaogang Wang$^{1}$  
	~~\quad Hongsheng Li$^{1}$	\vspace{0.2cm}\\
$^1$CUHK-SenseTime Joint Laboratory, The Chinese University of Hong Kong \\
$^2$SenseTime Research ~~\quad  
$^3$NLPR, CASIA ~~\quad
$^4$CSE, CUHK\\
}
\maketitle
\let\thefootnote\relax\footnotetext{
	E-mail: \{ssshi, hsli\}@ee.cuhk.edu.hk
}
\vspace{-2mm}

\begin{abstract}
We present a novel and high-performance 3D object detection framework, named PointVoxel-RCNN (PV-RCNN), for accurate 3D object detection from point clouds. Our proposed method deeply integrates both 3D voxel Convolutional Neural Network (CNN) and PointNet-based set abstraction to learn more discriminative point cloud features. 
It takes advantages of efficient learning and high-quality proposals of the 3D voxel CNN and the flexible receptive fields of the PointNet-based networks.
Specifically, the proposed framework summarizes the 3D scene with a 3D voxel CNN into a small set of keypoints via a novel voxel set abstraction module to save follow-up computations and also to encode representative scene features. Given the high-quality 3D proposals generated by the voxel CNN, the RoI-grid pooling is proposed to abstract proposal-specific features from the keypoints to the RoI-grid points via keypoint set abstraction with multiple receptive fields. Compared with conventional pooling operations, the RoI-grid feature points encode much richer context information for accurately estimating object confidences and locations. Extensive experiments on both the KITTI dataset and the Waymo Open dataset show that our proposed PV-RCNN surpasses state-of-the-art 3D detection methods with remarkable margins by using only point clouds.
Code is available at \url{https://github.com/open-mmlab/OpenPCDet}. 

\end{abstract}

\vspace{-4mm}
\section{Introduction}
3D object detection has been receiving increasing attention from both industry and academia thanks to its wide applications in various fields such as autonomous driving and robotics. 
LiDAR sensors are widely adopted in autonomous driving vehicles and robots for capturing 3D scene information as sparse and irregular point clouds, which provide vital cues for 3D scene perception and understanding.
In this paper, we propose to achieve high performance 3D object detection by designing novel 
point-voxel integrated networks
to learn better 3D features from irregular point clouds.

Most existing 3D detection methods could be classified into two categories in terms of point cloud representations, \ie, the grid-based methods and the point-based methods. The grid-based methods generally transform the irregular point clouds to regular representations such as 3D voxels \cite{song2016deep,zhou2018voxelnet,yan2018second,Chen2019fastpointrcnn,shi2019part} or 2D bird-view maps \cite{Chen2017CVPR,ku2018joint,yang2018pixor,Liang2018ECCV,Yang2018CoRL,lang2018pointpillars,Liang2019CVPR}, which could be efficiently processed by 3D or 2D Convolutional Neural Networks (CNN) to learn point features for 3D detection. Powered by the pioneer work, PointNet and its variants \cite{qi2017pointnet,qi2017pointnet++}, the point-based methods \cite{qi2017frustum,shi2019pointrcnn,wang2019frustum,std2019yang} directly extract discriminative features from raw point clouds for 3D detection. 
Generally, the grid-based methods are more computationally efficient but the inevitable information loss degrades the fine-grained localization accuracy, while the point-based methods have higher computation cost but could easily achieve larger receptive field by the point set abstraction \cite{qi2017pointnet++}. However, we show that a unified framework could integrate the best of the two types of methods, and surpass the prior state-of-the-art 3D detection methods with remarkable margins.

\begin{figure}[t]
	\begin{center}
			\includegraphics[width=1.0\linewidth]{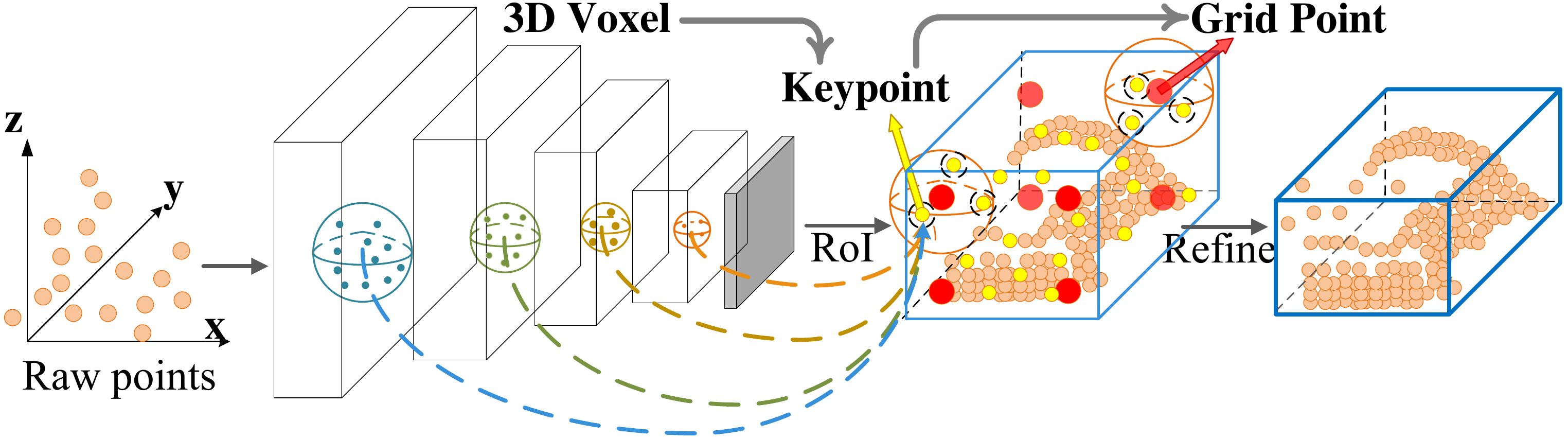}
	\end{center}
		\vspace{-0.3cm}
	\caption{Our proposed PV-RCNN framework deeply integrates both the voxel-based and the PointNet-based networks via a two-step strategy including the voxel-to-keypoint 3D scene encoding and the keypoint-to-grid RoI feature abstraction for improving the performance of 3D object detection.
	}
	\label{fig:teaser}
	\vspace{-0.3cm}
\end{figure}

We propose a novel 3D object detection framework, \textbf{PV-RCNN} (Illustrated in Fig.~\ref{fig:teaser}), which boosts the 3D detection performance by incorporating the advantages from both the \textbf{P}oint-based and \textbf{V}oxel-based feature learning methods. 
The principle of PV-RCNN lies in the fact that the voxel-based operation efficiently encodes multi-scale feature representations and can generate high-quality 3D proposals, while the PointNet-based set abstraction operation preserves accurate location information with flexible receptive fields. We argue that the integration of these two types of feature learning frameworks can help learn more discriminative features for accurate fine-grained box refinement.

The main challenge would be how to effectively combine the two types of feature learning schemes, specifically the 3D voxel CNN with sparse convolutions \cite{SubmanifoldSparseConvNet, 3DSemanticSegmentationWithSubmanifoldSparseConvNet} and the PointNet-based set abstraction \cite{qi2017pointnet++}, into a unified framework. 
An intuitive solution would be uniformly sampling several grid points within each 3D proposal, and adopt the set abstraction to aggregate 3D voxel-wise features surrounding these grid points for proposal refinement. However, this strategy is highly memory-intensive since both the number of voxels and the number of grid points could be quite large to achieve satisfactory performance. 

Therefore, to better integrate these two types of point cloud feature learning networks, we propose a two-step strategy with the first voxel-to-keypoint scene encoding step and the second keypoint-to-grid RoI feature abstraction step. 
Specifically, a voxel CNN with 3D sparse convolution is adopted for voxel-wise feature learning and accurate proposal generation.
To mitigate the above mentioned issue of requiring too many voxels for encoding the whole scene, a small set of keypoints are selected by the furtherest point sampling (FPS) to summarize the overall 3D information from the voxel-wise features. 
The features of each keypoint is aggregated by 
grouping the neighboring voxel-wise features via PointNet-based set abstraction
for summarizing multi-scale point cloud information.
In this way, the overall scene can be effectively and efficiently encoded by a small number of keypoints with associated multi-scale features.

For the second keypoint-to-grid RoI feature abstraction step, 
given each box proposal with its grid point locations, a RoI-grid pooling module is proposed, where a keypoint set abstraction layer with multiple radii is adopted for each grid point to aggregate the features from the keypoints with multi-scale context. All grid points' aggregated features can then be jointly used for the succeeding proposal refinement. Our proposed PV-RCNN effectively takes advantages of both point-based and voxel-based networks to encode discriminative features at each box proposal for accurate confidence prediction and fine-grained box refinement.

Our contributions can be summarized into four-fold. 
(1) We propose PV-RCNN framework which effectively takes advantages of both the voxel-based and point-based methods for 3D point-cloud feature learning, leading to improved performance of 3D object detection with manageable memory consumption.
(2) We propose the voxel-to-keypoint scene encoding scheme, 
which encodes multi-scale voxel features of the whole scene 
to a small set of keypoints by the voxel set abstraction layer. These keypoint features not only preserve accurate location but also encode rich scene context, which boosts the 3D detection performance significantly. 
(3) We propose a multi-scale RoI feature abstraction layer for grid points in each proposal, which aggregates richer context information from the scene with multiple receptive fields for accurate box refinement and confidence prediction.
(4) Our proposed method PV-RCNN outperforms all previous methods with remarkable margins and ranks $1^{st}$ on the highly competitive KITTI 3D detection benchmark \cite{kitti_leaderboard}, ans also surpasses previous methods on the large-scale Waymo Open dataset with a large margin.

\section{Related Work}

\noindent
\textbf{3D Object Detection with Grid-based Methods.}~
To tackle the irregular data format of point clouds, most existing works project the point clouds to regular grids to be processed by 2D or 3D CNN. The pioneer work MV3D \cite{Chen2017CVPR} projects the point clouds to 2D bird view grids and places lots of predefined 3D anchors for generating 3D bounding boxes, and the following works \cite{ku2018joint,Liang2018ECCV,Liang2019CVPR} develop better strategies for multi-sensor fusion while \cite{yang2018pixor,Yang2018CoRL,lang2018pointpillars} propose more efficient frameworks with bird view representation. 
Some other works \cite{song2016deep,zhou2018voxelnet} divide the point clouds into 3D voxels to be processed by 3D CNN, and 3D sparse convolution \cite{3DSemanticSegmentationWithSubmanifoldSparseConvNet} is introduced \cite{yan2018second} for efficient 3D voxel processing.
\cite{wang2019voxelFPN,zhu2019class} utilizes multiple detection heads while \cite{shi2019part} explores the object part locations for improving the performance.
These grid-based methods are generally efficient for accurate 3D proposal generation but the receptive fields are constraint by the kernel size of 2D/3D convolutions.

\noindent
\textbf{3D Object Detection with Point-based Methods.}~
F-PointNet \cite{qi2017frustum} first proposes to apply PointNet \cite{qi2017pointnet,qi2017pointnet++} for 3D detection from the cropped point clouds based on the 2D image bounding boxes. PointRCNN \cite{shi2019pointrcnn} generates 3D proposals directly from the whole point clouds instead of 2D images for 3D detection with point clouds only, and the following work STD \cite{std2019yang} proposes the sparse to dense strategy for better proposal refinement. \cite{qi2019deep} proposes the hough voting strategy for better object feature grouping.
These point-based methods are mostly based on the PointNet series, especially the set abstraction operation \cite{qi2017pointnet++}, which enables flexible receptive fields for point cloud feature learning. 

\noindent
\textbf{Representation Learning on Point Clouds.}~
Recently representation learning on point clouds has drawn lots of attention for improving the performance of point cloud classification and segmentation \cite{qi2017pointnet,qi2017pointnet++,zhou2018voxelnet,wang2019dynamic,huang2018recurrent,zhao2019pointweb,li2018pointcnn,su2018splatnet,wu2019pointconv,jaritz2019multi,thomas2019kpconv,choy20194d}. 
In terms of 3D detection, previous methods generally project the point clouds to regular bird view grids \cite{Chen2017CVPR,yang2018pixor} or 3D voxels \cite{zhou2018voxelnet,Chen2019fastpointrcnn} for processing point clouds with 2D/3D CNN. 3D sparse convolution \cite{SubmanifoldSparseConvNet,3DSemanticSegmentationWithSubmanifoldSparseConvNet} are adopted in \cite{yan2018second,shi2019part} to effectively learn sparse voxel-wise features from the point clouds. Qi \etal  \cite{qi2017pointnet,qi2017pointnet++} proposes the PointNet to directly learn point-wise features from the raw point clouds, where set abstraction operation enables flexible receptive fields by setting different search radii. \cite{liu2019point} combines both voxel-based CNN and point-based SharedMLP for efficient point cloud feature learning. 
In comparison, 
our proposed PV-RCNN takes advantages from both the voxel-based feature learning (\ie, 3D sparse convolution) and PointNet-based feature learning (\ie, set abstraction operation) to enable both high-quality 3D proposal generation and flexible receptive fields for improving the 3D detection performance.


\begin{figure*}
	\vspace{-5mm}
	\begin{center}
		\includegraphics[width=1.0\linewidth,height=6.2cm]{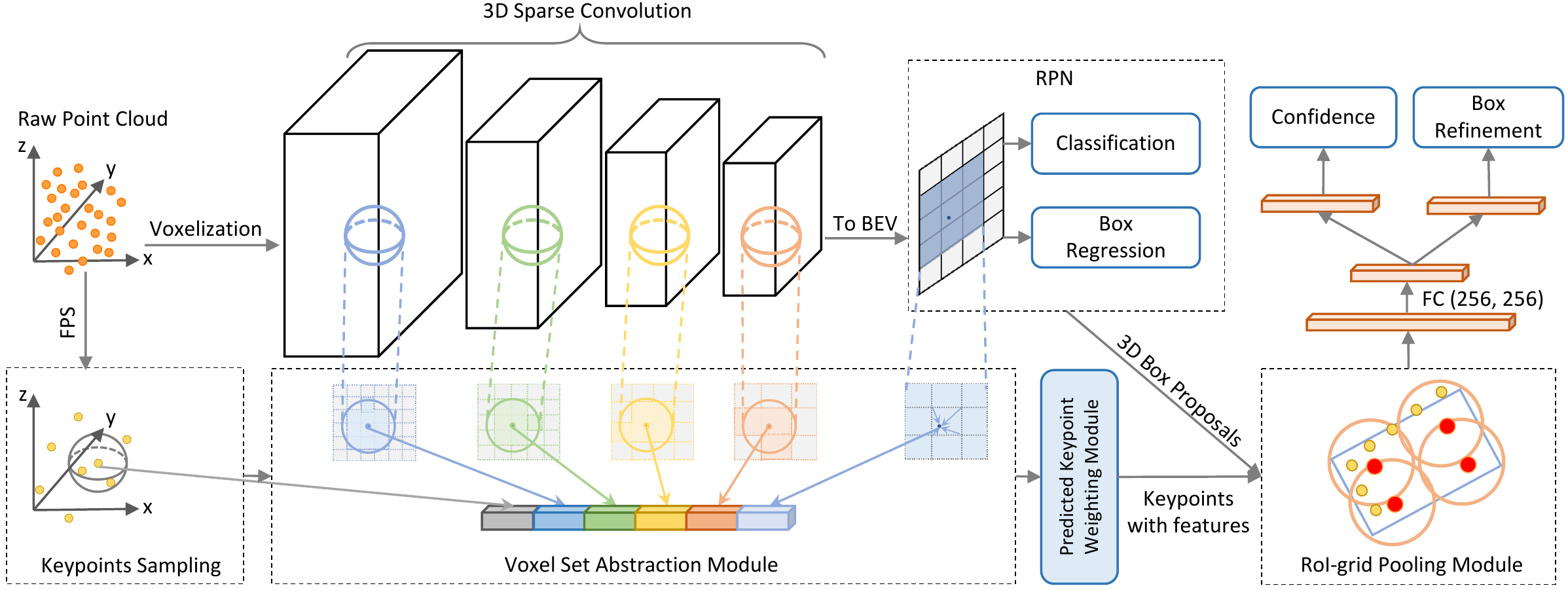}
	\end{center}
	\vspace{-0.38cm}
	\caption{The overall architecture of our proposed PV-RCNN. The raw point clouds are first voxelized to feed into the 3D sparse convolution based encoder to learn multi-scale semantic features and generate 3D object proposals. Then the learned voxel-wise feature volumes at multiple neural layers are summarized into a small set of key points via the novel voxel set abstraction module. Finally the keypoint features are aggregated to the RoI-grid points to learn proposal specific features for fine-grained proposal refinement and confidence prediction.}
	\label{fig:framework_total}
	\vspace{-0.4cm}
\end{figure*}

\section{PV-RCNN for Point Cloud Object Detection}

In this paper, we propose the PointVoxel-RCNN (PV-RCNN), which is a two-stage 3D detection framework aiming at more accurate 3D object detection from point clouds. 
State-of-the-art 3D detection approaches are based on either 3D voxel CNN with sparse convolution or PointNet-based networks as the backbone. Generally, the 3D voxel CNNs with sparse convolution are more efficient \cite{yan2018second,shi2019part} and are able to generate high-quality 3D object proposals, while the PointNet-based methods can capture more accurate contextual information with flexible receptive fields.

Our PV-RCNN deeply integrates the advantages of two types of networks. As illustrated in Fig.~\ref{fig:framework_total}, the PV-RCNN consists of a 3D voxel CNN with sparse convolution as the backbone for efficient feature encoding and proposal generation. Given each 3D object proposal, to effectively pool its corresponding features from the scene, we propose two novel operations: the voxel-to-keypoint scene encoding, which summarizes all the voxels of the overall scene feature volumes into a small number of feature keypoints, and the point-to-grid RoI feature abstraction, which effectively aggregates the scene keypoint features to RoI grids for proposal confidence prediction and location refinement.

\subsection{3D Voxel CNN for Efficient Feature Encoding and Proposal Generation}\label{sec:rpn}
Voxel CNN with 3D sparse convolution \cite{SubmanifoldSparseConvNet, 3DSemanticSegmentationWithSubmanifoldSparseConvNet,yan2018second,shi2019part} is a popular choice by state-of-the-art 3D detectors for efficiently converting the point clouds into sparse 3D feature volumes. Because of its high efficiency and accuracy, we adopt it as the backbone 
of our framework for feature encoding and 3D proposal generation.

\textit{3D voxel CNN.} The input points $\bf{P}$ are first divided into small voxels with spatial resolution of $L\times W \times H$, where the features of the non-empty voxels are directly calculated as the mean of point-wise features of all inside points. The commonly used features are the 3D coordinates and reflectance intensities. 
The network utilizes a series of $3\times 3 \times 3$ 3D sparse convolution to gradually convert the point clouds into feature volumes with $1\times, 2\times$, $4\times$, $8\times$ downsampled sizes. 
Such sparse feature volumes at each level could be viewed as a set of voxel-wise feature vectors.

\textit{3D proposal generation.} By converting the encoded $8\times$ downsampled 3D feature volumes into 2D bird-view feature maps, high-quality 3D proposals are generated following the anchor-based approaches \cite{yan2018second,lang2018pointpillars}. Specifically, we stack the 3D feature volume along the $Z$ axis to obtain the $\frac{L}{8} \times \frac{W}{8}$ bird-view feature maps. 
Each class has $2\times\frac{L}{8} \times \frac{W}{8}$ 3D anchor boxes
which 
adopt the average 3D object sizes of this class, and two anchors of $0^\circ, 90^\circ$ orientations are evaluated for each pixel of the bird-view feature maps. 
As shown in Table~\ref{tab:val_recall}, the adopted 3D voxel CNN backbone with anchor-based scheme achieves higher recall performance than the PointNet-based approaches \cite{shi2019pointrcnn,std2019yang}.

\textit{Discussions.}~
State-of-the-art detectors mostly adopt two-stage frameworks. They require pooling RoI specific features from the resulting 3D feature volumes or 2D maps for further proposal refinement. 
However, these 3D feature volumes from the 3D voxel CNN have major limitations in the following aspects. (i) These feature volumes
are generally of low spatial resolution as they are downsampled by up to 8 times, which hinders accurate localization of objects in the input scene. (ii) Even if one can upsample to obtain feature volumes/maps of larger spatial sizes, they are generally still quite sparse. The commonly used trilinear or bilinear interpolation in the RoIPooling/RoIAlign operations can only extract features from very small neighborhoods (i.e., 4 and 8 nearest neighbors for bilinear and trilinear interpolation respectively). The conventional pooling approaches would therefore obtain features with mostly zeros and waste much computation and memory for stage-2 refinement.

On the other hand, the set abstraction operation proposed in the variants of PointNet \cite{qi2017pointnet,qi2017pointnet++} has shown the strong capability of encoding feature points from a neighborhood of an arbitrary size. We therefore propose to integrate a 3D voxel CNN with a series of set abstraction operations for conducting accurate and robust stage-2 proposal refinement.

A naive solution of using the set abstraction operation for pooling the scene feature voxels would be directly aggregating 
the multi-scale feature volume in a scene to the RoI grids. However, this intuitive strategy simply occupies much memory and is inefficient to be used in practice. For instance, a common scene from the KITTI dataset might result in $18,000$ voxels in the $4\times$ downsampled feature volumes. If one uses $100$ box proposal for each scene and each box proposal has $3\times 3 \times 3$ grids. The $2,700 \times 18,000$ pairwise distances and feature aggregations cannot be efficiently computed, even after distance thresholding.

To tackle this issue, we propose a two-step approach to first encode voxels at different neural layers of the entire scene into a small number of keypoints and then aggregate keypoint features to RoI grids for box proposal refinement.

\subsection{Voxel-to-keypoint Scene Encoding via Voxel Set Abstraction}\label{voxel_to_point}

Our proposed framework first aggregates the voxels at the multiple neural layers representing the entire scene into a small number of keypoints, which serve as a bridge between the 3D voxel CNN feature encoder and the proposal refinement network. 

\noindent
\textbf{Keypoints Sampling.}~
Specifically, we adopt the Furthest-Point-Sampling (FPS) algorithm to sample a small number of $n$ keypoints $\mathcal{K}=\left\{p_1, \cdots, p_n\right\}$ from the point clouds $\bf{P}$, where $n=2,048$ for the KITTI dataset and $n=4,096$ for the Waymo dataset. Such a strategy encourages that the keypoints are uniformly distributed around non-empty voxels and can be representative to the overall scene.

\noindent
\textbf{Voxel Set Abstraction Module.}~
We propose the \textit{Voxel Set Abstraction} (VSA) module to encode the multi-scale semantic features from the 3D CNN feature volumes to the keypoints. 
The set abstraction operation proposed by \cite{qi2017pointnet++} is adopted for the aggregation of voxel-wise feature volumes. The surrounding points of keypoints are now regular voxels with multi-scale semantic features encoded by the 3D voxel CNN from the multiple levels, 
instead of the neighboring raw points with features learned from PointNet. 

Specifically, denote \small $\mathcal{F}^{(l_k)}=\{f_1^{(l_k)}, \cdots, f_{N_k}^{(l_k)}\}$ \normalsize as the set of voxel-wise feature vectors in the $k$-th level of 3D voxel CNN, \small $\mathcal{V}^{(l_k)}=\{v_1^{(l_k)}, \cdots, v_{N_k}^{(l_k)}\}$ \normalsize as their 3D coordinates calculated by the voxel indices and actual voxel sizes of the $k$-th level, 
where $N_k$ is the number of non-empty voxels in the $k$-th level. 
For each keypoint $p_i$, we first identify its neighboring non-empty voxels at the $k$-th level within a radius $r_k$ to retrieve the set of voxel-wise feature vectors as 
{
\vspace{-5mm}
\small
\begin{align}
S_i^{(l_k)} =\left\lbrace \left[f_j^{(l_k)}; v_j^{(l_k)} - p_i\right]^T \;\middle|\;
\begin{tabular}{@{}l@{}}
$\left\Vert v_j^{(l_k)} - p_i \right\Vert^2 < r_k$, \\
$\forall v_j^{(l_k)} \in \mathcal{V}^{(l_k)}$, \\
$\forall f_j^{(l_k)} \in \mathcal{F}^{(l_k)}$ \\
\end{tabular}
\right\rbrace,
\end{align}
}
\vspace{-3mm}

\noindent 
where we concatenate the local relative coordinates $v_j^{(l_k)}$ $- p_i$ to indicate the relative location of semantic voxel feature $f_j^{(l_k)}$.
The voxel-wise features within the neighboring voxel set $S_i^{(l_k)}$ of $p_i$ are then transformed by a PointNet-block \cite{qi2017pointnet} to generate the feature for the key point $p_i$ as
\vspace{-4mm}
\begin{align}\label{pointnet}
f_i^{(pv_k)} = \max \left\{G\left(\mathcal{M}\left(S_i^{(l_k)}\right)\right)\right\},
\end{align}
\vspace{-4mm}

\noindent 
where $\mathcal{M}(\cdot)$ denotes randomly sampling at most $T_k$ voxels from the neighboring set $S_i^{(l_k)}$ for saving computations, $G(\cdot)$ denotes a multi-layer perceptron network to encode the voxel-wise features and relative locations. Although the number of neighboring voxels varies across different keypoints, the along-channel max-pooling operation $\max(\cdot)$ maps the diverse number of neighboring voxel feature vectors to a feature vector $f_i^{(pv_k)}$ for the key point $p_i$.
Generally, we also set multiple radii $r_k$ at the $k$-th level to aggregate local voxel-wise features with different receptive fields for capturing richer multi-scale contextual information.

The above voxel set abstraction is performed at different levels of the 3D voxel CNN, and the aggregated features from different levels can be concatenated to generate the multi-scale semantic feature for the key point $p_i$
{
\vspace{-2mm}
\small 
\begin{align}\label{eq:keypointfeature0}
f_i^{(pv)} = \left[f_i^{(pv_1)}, f_i^{(pv_2)}, f_i^{(pv_3)}, f_i^{(pv_4)} \right], \text{ for } i = 1, \cdots, n,
\end{align}
}
\vspace{-4mm}
\normalsize 

\noindent
where the generated feature $f_i^{(pv)}$ incorporates both the 3D voxel CNN-based feature learning from voxel-wise feature $f_j^{(l_k)}$ and the PointNet-based feature learning from voxel set abstraction as Eq.~\eqref{pointnet}. Besides, the 3D coordinate of $p_i$ also preserves accurate location information.

\noindent
\textbf{Extended VSA Module.} We extend the VSA module by further enriching the keypoint features from the raw point clouds $\bf{P}$ and the $8\times$ downsampled 2D bird-view feature maps (as described in Sec. \ref{sec:rpn}), where the raw point clouds partially make up the quantization loss of the initial point-cloud voxelization while the 2D bird-view maps have larger receptive fields along the $Z$ axis.
The raw point-cloud feature $f_i^{(raw)}$ is also aggregated as in Eq.~\eqref{pointnet}. For the bird view feature maps, we project the keypoint $p_i$ to the 2D bird-view coordinate system, and utilize bilinear interpolation to obtain the features $f_i^{(bev)}$ from the bird-view feature maps. Hence, the keypoint feature for $p_i$ is further enriched by concatenating all its associated features
\vspace{-2mm}
\begin{align}\label{eq:keypointfeatures1}
f_i^{(p)} = \left[f_i^{(pv)}, f_i^{(raw)}, f_i^{(bev)} \right], 
\text{ for } i = 1, \cdots, n,
\end{align}
\vspace{-6mm}

\noindent 
which have the strong capability of preserving 3D structural information of the entire scene and can also boost the final detection performance by large margins. 

\begin{figure}
	\vspace{-3mm}
	\begin{center}
		\includegraphics[width=0.9\linewidth,height=4.0cm]{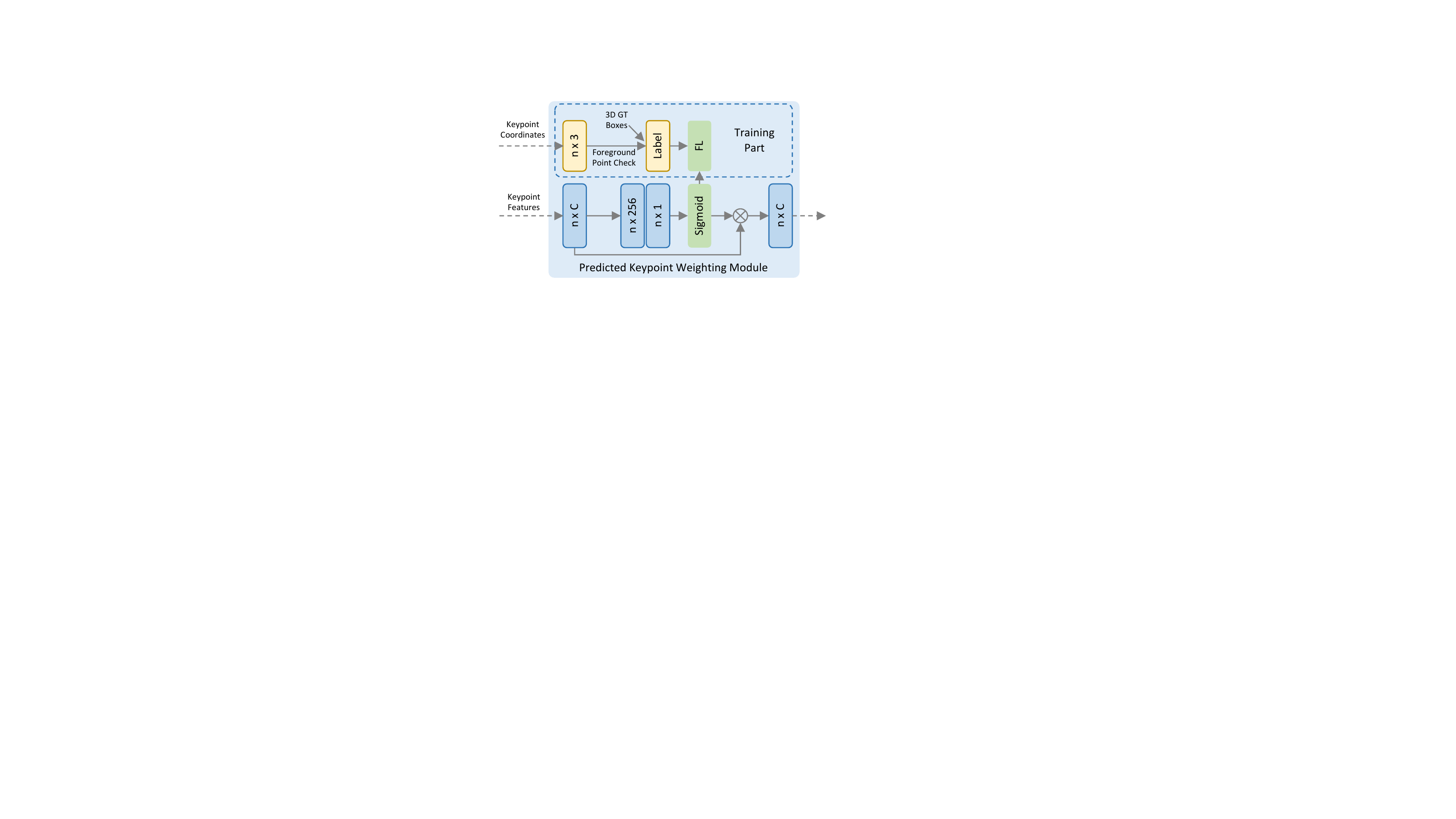}
	\end{center}
	\vspace{-0.3cm}
	\caption{Illustration of Predicted Keypoint Weighting module.}
	\label{fig:KPA}
	\vspace{-0.5cm}
\end{figure}

\noindent
\textbf{Predicted Keypoint Weighting.}~
After the overall scene is encoded by a small number of keypoints, they would be further utilized by the succeeding stage for conducting proposal refinement. 
The keypoints are chosen by the Further Point Sampling strategy and some of them might only represent the background regions. Intuitively, keypoints belonging to the foreground objects should contribute more to the accurate refinement of the proposals, while the ones from the background regions should contribute less.

Hence, we propose a \textit{Predicted Keypoint Weighting} (PKW) module (see Fig.~\ref{fig:KPA}) to re-weight the keypoint features with extra supervisions from point-cloud segmentation. The segmentation labels can be directly generated by the 3D detection box annotations, \ie by checking whether each key point is inside or outside of a ground-truth 3D box since the 3D objects in autonomous driving scenes are naturally separated in 3D space. The predicted feature weighting for each keypoint's feature $\tilde{f}_i^{(p)}$ can be formulated as
\vspace{-5mm}
\begin{align}\label{eq:pkw}
\tilde{f}_i^{(p)} = \mathcal{A}(f_i^{(p)}) \cdot f_i^{(p)},
\end{align}
\vspace{-5mm}

\noindent
where $\mathcal{A}(\cdot)$ is a three-layer MLP network with a sigmoid function to predict foreground confidence between $[0,1]$. 
The PKW module is trained by focal loss \cite{lin2018focal} with default hyper-parameters for handling the unbalanced number of foreground/background points in the training set.

\subsection{Keypoint-to-grid RoI Feature Abstraction for Proposal Refinement}

In the previous step, the whole scene is summarized into a small number of keypoints with multi-scale semantic features. Given each 3D proposal (RoI) generated by the 3D voxel CNN,
the features of each RoI need to be aggregated from the keypoint features $\tilde{\mathcal{F}}=\{\tilde{f}^{(p)}_1, \cdots, \tilde{f}^{(p)}_n\}$ for 
accurate and robust proposal refinement.
We propose the keypoint-to-grid RoI feature abstraction based on the set abstraction operation for multi-scale RoI feature encoding.

\begin{figure}[t]
	\vspace{-4mm}
	\begin{center}
		\includegraphics[width=0.95\linewidth,height=4.cm]{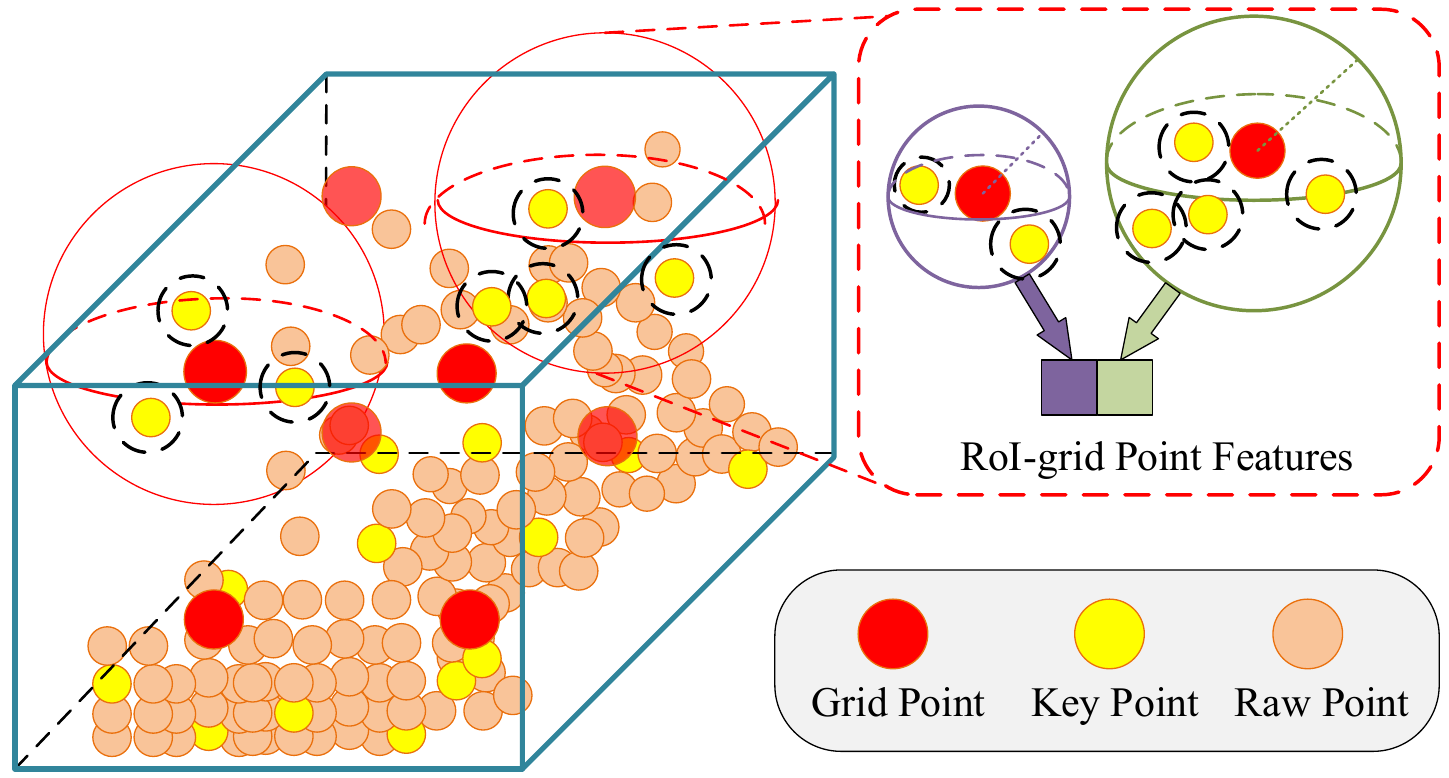}
	\end{center}
	\vspace{-0.4cm}
	\caption{Illustration of RoI-grid pooling module. Rich context information of each 3D RoI is aggregated by the set abstraction operation with multiple receptive fields.}
	\label{fig:roipool}
	\vspace{-0.3cm}
\end{figure}

\noindent
\textbf{RoI-grid Pooling via Set Abstraction.}~
Given each 3D RoI, as shown in Fig.~\ref{fig:roipool}, we propose the RoI-grid pooling module to aggregate the keypoint features to the RoI-grid points with multiple receptive fields. We uniformly sample $6\times 6\times 6$ grid points within each 3D proposal, which are denoted as $\mathcal{G}=\{g_1, \cdots, g_{216}\}$. 
The set abstraction operation is adopted to aggregate the features of grid points from the keypoint features. 
Specifically, we firstly identify the neighboring keypoints of grid point $g_i$ within a radius $\tilde{r}$ as 
{
\vspace{-1mm}
\small
\begin{align}
\tilde{\Psi} =\left\lbrace \left[\tilde{f}_j^{(p)}; p_j - g_i\right]^T \;\middle|\;
\begin{tabular}{@{}l@{}}
$\left\Vert p_j - g_i \right\Vert^2 < \tilde{r}$, \\
$\forall p_j \in \mathcal{K}$, 
$\forall \tilde{f}_j^{(p)} \in \tilde{\mathcal{F}}$ \\
\end{tabular}
\right\rbrace,
\end{align}
} 
\vspace{-3mm}

\noindent where $p_j-g_i$ is appended to indicate the local relative location of features $\tilde{f}^{(p)}_j$ from keypoint $p_j$. Then a PointNet-block \cite{qi2017pointnet} is adopted to aggregate the neighboring keypoint feature set $\tilde{\Psi}$ to generate the feature for grid point $g_i$ as 
\vspace{-2mm}
\small 
\begin{align}\label{pointnet2}
\tilde{f}_i^{(g)} = \max \left\{G\left(\mathcal{M}\left(\tilde{\Psi}\right)\right)\right\},
\end{align}
\normalsize 
where $\mathcal{M}(\cdot)$ and $G(\cdot)$ are defined as the same in Eq.~\eqref{pointnet}. 
We set multiple radii $\tilde{r}$ and aggregate keypoint features with different receptive fields, which are concatenated together for capturing richer multi-scale contextual information. 

After obtaining each grid's aggregated features from its surrounding keypoints, all RoI-grid features of the same RoI can be vectorized and transformed by a two-layer MLP with $256$ feature dimensions to represent the overall proposal. 

Compared with the point cloud 3D RoI pooling operations in previous works \cite{shi2019pointrcnn,std2019yang,shi2019part}, our proposed RoI-grid pooling operation targeting the keypoints is able to capture much richer contextual information with flexible receptive fields, where the receptive fields are even beyond the RoI boundaries for capturing the surrounding keypoint features outside the 3D RoI, while the previous state-of-the-art methods either simply average all point-wise features within the proposal as the RoI feature \cite{shi2019pointrcnn}, 
or pool many uninformative zeros as the RoI features \cite{shi2019part,std2019yang}. 

\noindent
\textbf{3D Proposal Refinement and Confidence Prediction.}~
Given the RoI feature of each box proposal, the proposal refinement network learns to predict the size and location (\ie, center, size and orientation) residuals relative to the input 3D proposal. The refinement network adopts a 2-layer MLP and has two branches for confidence prediction and box refinement respectively.

For the confidence prediction branch, 
we follow \cite{li2019gs3d,jiang2018acquisition,shi2019part} to adopt the 3D Intersection-over-Union (IoU) between the 3D RoIs and their corresponding ground-truth boxes as the training targets. For the $k$-th 3D RoI, its confidence training target $y_k$ is normalized to be between $[0,1]$ as
\vspace{-2mm}
\begin{align}\label{eq:iou}
	y_k = \min \left(1,~ \max \left(0,~ 2\text{IoU}_k - 0.5\right)\right),
\end{align}
\vspace{-5mm}

\noindent
where $\text{IoU}_k$ is the IoU of the $k$-th RoI w.r.t. its ground-truth box. 
Our confidence branch is then trained to minimize the cross-entropy loss on predicting the confidence targets,
\vspace{-2mm}
\begin{align}\label{loss:bce}
L_{\textrm{iou}} = -y_k\log(\tilde{y}_k) - (1 - y_k)\log(1 - \tilde{y}_k),
\end{align}
\normalsize 
\vspace{-6mm}

\noindent
where $\tilde{y}_k$ is the predicted score by the network.
Our experiments in Table~\ref{tab:ab_study} show that this quality-aware confidence prediction strategy achieves better performance than the traditional classification targets. 

The box regression targets of the box refinement branch are encoded by the traditional residual-based method as in \cite{yan2018second,shi2019part} and are optimized by smooth-L1 loss function.

\subsection{Training losses}
The proposed PV-RCNN framework is trained end-to-end with the region proposal loss $L_{\text{rpn}}$, keypoint segmentation loss $L_{\text{seg}}$ and the proposal refinement loss $L_{\text{rcnn}}$. (1) We adopt the same region proposal loss $L_{\text{rpn}}$ with \cite{yan2018second} as 
\vspace{-1mm}
\small 
\begin{align}
L_{\textrm{rpn}} =L_{\text{cls}} + \beta\sum_{\text{r} \in \{x, y, z, l, h, w, \theta\}} \mathcal{L}_{\text{smooth-L1}}(\widehat{\Delta \text{r}^{a}}, \Delta \text{r}^{a}),
\end{align}
\normalsize
\vspace{-3mm}

\noindent
where the anchor classification loss $L_\text{cls}$ is calculated with focal loss \cite{lin2018focal} with default hyper-parameters and smooth-L1 loss is utilized for anchor box regression with the predicted residual $\widehat{\Delta \text{r}^{a}}$ and the regression target $\Delta \text{r}^{a}$.
(2) The keypoint segmentation loss $L_{\textrm{seg}}$ is also calculated by the focal loss as mentioned in Sec.~\ref{voxel_to_point}. (3) The proposal refinement loss $L_{\text{rcnn}}$ includes the IoU-guided confidence prediction loss $L_{\text{iou}}$ and the box refinement loss as  
\vspace{-2mm}
\small 
\begin{align}
L_{\textrm{rcnn}} =L_{\text{iou}} + \sum_{\text{r} \in \{x, y, z, l, h, w, \theta\}} \mathcal{L}_{\text{smooth-L1}}(\widehat{\Delta \text{r}^{p}}, \Delta \text{r}^{p}),
\end{align}
\normalsize
\vspace{-4mm}

\noindent
where $\widehat{\Delta \text{r}^{p}}$ is the predicted box residual and $\Delta \text{r}^{p}$ is the proposal regression target which are encoded same with $\Delta \text{r}^{a}$. 

The overall training loss are then the sum of these three losses with equal loss weights. Further training loss details are provided in the supplementary file.

\section{Experiments}
In this section, we introduce the implementation details of our PV-RCNN framework (Sec.~\ref{sec:imple}) and compare 
with previous state-of-the-art methods on both the highly competitive KITTI dataset \cite{Geiger2012CVPR} (Sec.~\ref{sec:kitti}) and the newly introduced large-scale Waymo Open Dataset \cite{ngiam2019starnet,zhou2019end} (Sec.~\ref{sec:waymo}). In Sec. \ref{sec:ab}, we conduct extensive ablation studies to investigate each component of PV-RCNN to validate our design.

\subsection{Experimental Setup}\label{sec:imple}
\noindent
\textbf{Datasets.}~
\textit{KITTI Dataset} \cite{Geiger2012CVPR} is one of the most popular dataset of 3D detection for autonomous driving. There are $7,481$ training samples and $7,518$ test samples, where the training samples are generally divided into the \textit{train} split ($3,712$ samples) and the \textit{val} split ($3,769$ samples). We compare PV-RCNN with state-of-the-art methods on both the \textit{val} split and the \textit{test} split on the online learderboard.

\textit{Waymo Open Dataset} is a recently released and currently the largest dataset of 3D detection for autonomous driving. There are totally $798$ training sequences with around $158,361$ LiDAR samples, and $202$ validation sequences with $40,077$ LiDAR samples. It annotated the objects in the full $360^{\circ}$  field instead of $90^{\circ}$ in KITTI dataset. We evaluate our model on this large-scale dataset to further validate the effectiveness of our proposed method.

\noindent
\textbf{Network Architecture.}~
As shown in Fig.~\ref{fig:framework_total}, the 3D voxel CNN has four levels with feature dimensions $16, 32, 64, 64$, respectively. Their two neighboring radii $r_k$ of each level in the VSA module are set as $(0.4\text{m}, 0.8\text{m})$, $(0.8\text{m},1.2\text{m})$, $(1.2\text{m}, 2.4\text{m})$, $(2.4\text{m}, 4.8\text{m})$, and 
the neighborhood radii of set abstraction for raw points are $(0.4\text{m}, 0.8\text{m})$. For the proposed RoI-grid pooling operation, we uniformly sample $6\times6\times6$ grid points in each 3D proposal and the two neighboring radii $\tilde{r}$ of each grid point are $(0.8\text{m}, 1.6\text{m})$. 

For the KITTI dataset, the detection range is within $[0, 70.4]m$ for the $X$ axis, $[-40,40]m$ for the $Y$ axis and $[-3, 1]m$ for the $Z$ axis, which is voxelized with the voxel size $(0.05m, 0.05m, 0.1m)$ in each axis. For the Waymo Open dataset, the detection range is $[-75.2, 75.2]m$ for the $X$ and $Y$ axes and $[-2, 4]m$ for the $Z$ axis, and we set the voxel size to $(0.1m, 0.1m, 0.15m)$.

\begin{table*}
	\small 
	\vspace{-2mm}
	\begin{center}
		\scalebox{0.83}[0.8]{
			\setlength\tabcolsep{5pt}
			\begin{tabular}{c|c|c||ccc|ccc|ccl|ccl}
				\hline
				\multirow{2}{*}{Method} & 
				\multirow{2}{*}{Reference} & 
				\multirow{2}{*}{Modality} & 			
				\multicolumn{3}{c|}{~~Car - 3D Detection ~~} & \multicolumn{3}{c|}{~Car - BEV Detection~} & \multicolumn{3}{c|}{~Cyclist - 3D Detection~} & \multicolumn{3}{c}{Cyclist - BEV Detection}\\
				&&&Easy & Mod. & Hard & Easy & Mod. & Hard & Easy & Mod. & Hard & Easy & Mod. & Hard\\
				\hline
				MV3D \cite{Chen2017CVPR} & CVPR 2017 & RGB + LiDAR & 74.97 & 63.63 & 54.00 & 86.62 & 78.93 & 69.80 & - & - & - & - & - & -\\
				ContFuse \cite{Liang2018ECCV} & ECCV 2018 & RGB + LiDAR & 83.68 & 68.78 & 61.67 & 94.07 & 85.35 & 75.88 & - & - & - & - & - & -\\ 
				AVOD-FPN \cite{ku2018joint} & IROS 2018 & RGB + LiDAR & 83.07 & 71.76 & 65.73 & 90.99 & 84.82 & 79.62 & 63.76 & 50.55 & 44.93 & 69.39 & 57.12 & 51.09 \\			
				F-PointNet \cite{qi2017frustum} & CVPR 2018 & RGB + LiDAR & 82.19 & 69.79 & 60.59 & 91.17 & 84.67 & 74.77 & 72.27 & 56.12 & 49.01 & 77.26 & 61.37 & 53.78 \\
				UberATG-MMF \cite{Liang2019CVPR} & CVPR 2019 & RGB + LiDAR & 88.40 & 77.43 & 70.22 & 93.67 & 88.21 & 81.99 & - & - & - & - & - & -\\	
				\hline 		
				SECOND \cite{yan2018second} & Sensors 2018 & LiDAR only & 83.34 & 72.55 & 65.82 & 89.39 & 83.77 & 78.59 & 71.33 & 52.08 & 45.83 & 76.50 & 56.05 & 49.45 \\
				PointPillars \cite{lang2018pointpillars} & CVPR 2019 & LiDAR only & 82.58 & 74.31 & 68.99 & 90.07 & 86.56 & 82.81 & 77.10 & 58.65 & 51.92 & 79.90 & 62.73 & 55.58 \\
				PointRCNN \cite{shi2019pointrcnn} & CVPR 2019 & LiDAR only & 86.96 & 75.64 & 70.70  & 92.13 & 87.39 & 82.72 & 74.96 & 58.82 & 52.53 & \textbf{82.56} & 67.24 & 60.28 \\
				3D IoU Loss \cite{zhou2019} & 3DV 2019 & LiDAR only & 86.16 & 76.50 & 71.39 & 91.36 & 86.22 & 81.20 & - & - & - & - & - &- \\
				Fast Point R-CNN \cite{Chen2019fastpointrcnn} & ICCV 2019 & LiDAR only & 85.29 & 77.40 & 70.24 & 90.87 & 87.84 & 80.52 & - & - & - & - & - &- \\
				STD \cite{std2019yang} & ICCV 2019 & LiDAR only & 87.95 & 79.71 & 75.09 & 94.74 & 89.19 & \textbf{86.42} & \textbf{78.69} & 61.59 & 55.30 & 81.36 & 67.23 & 59.35 \\
				Patches \cite{lehner2019patch} & Arxiv 2019 & LiDAR only & 88.67 & 77.20 & 71.82 & 92.72 & 88.39 & 83.19 & - & - & - & - & - &- \\
				Part-A2-Net \cite{shi2019part} & TPAMI 2020 & LiDAR only & 87.81 & 78.49 & 73.51 & 91.70 & 87.79 & 84.61 & - & - & - & - & - &- \\
				\hline 
				PV-RCNN (Ours) & - & LiDAR only & \textbf{90.25} & \textbf{81.43} & \textbf{76.82} & \textbf{94.98} & \textbf{90.65} & 86.14 & 78.60 & \textbf{63.71} & \textbf{57.65} & 82.49 & \textbf{68.89} & \textbf{62.41} \\
				\rowcolor{LightCyan}
				\textit{Improvement} & - & - & \textit{+1.58} & \textit{+1.72} & \textit{+1.73} & \textit{+0.24} & \textit{+1.46} & \textit{-0.28} & \textit{-0.06} & \textit{+2.12} & \textit{+2.35} & \textit{-0.07} & \textit{+1.65} & \textit{+2.13} \\
				\hline
			\end{tabular}
		}
	\end{center}
\vspace{-0.2cm}
	\caption{Performance comparison on the KITTI \textit{test} set. The results are evaluated by the mean Average Precision with 40 recall positions.
	}
	\label{tab:test}
	\vspace{-0.2cm}
\end{table*}

\noindent
\textbf{Training and Inference Details.}~
Our PV-RCNN framework is trained from scratch in an end-to-end manner with the ADAM optimizer. 
For the KITTI dataset, we train the entire network with the batch size $24$, learning rate 0.01 for 80 epochs on 8 GTX 1080 Ti GPUs, which takes around 5 hours. 
For the Waymo Open Dataset, we train the entire network with batch size $64$, learning rate 0.01 for 30 epochs on 32 GTX 1080 Ti GPUs.
The cosine annealing learning rate strategy is adopted for the learning rate decay. 
For the proposal refinement stage, we randomly sample $128$ proposals with $1$:$1$ ratio for positive and negative proposals, where a proposal is considered as a positive proposal for box refinement branch if it has at least $0.55$ 3D IoU with the ground-truth boxes, otherwise it is treated as a negative proposal. 

During training, we utilize the widely adopted data augmentation strategy of 3D object detection, including random flipping along the $X$ axis, global scaling with a random scaling factor sampled from $[0.95, 1.05]$, global rotation around the $Z$ axis with a random angle sampled from $[-\frac{\pi}{4}, \frac{\pi}{4}]$. We also conduct the ground-truth sampling augmentation \cite{yan2018second} to randomly ``paste" some new ground-truth objects from other scenes to the current training scenes, for simulating objects in various environments. 

For inference, we keep the top-$100$ proposals generated from the 3D voxel CNN with a 3D IoU threshold of $0.7$ for non-maximum-suppression (NMS). These proposals are further refined in the proposal refinement stage with aggregated keypoint features. We finally use an NMS threshold of $0.01$ to remove the redundant boxes.

\begin{table}
	\small 
	\begin{center}
		\scalebox{0.85}[0.8]{
			\begin{tabular}{c|c|c|c}
				\hline
				Method & 
				Reference & 
				Modality& 3D mAP\\
				
				\hline
				MV3D \cite{Chen2017CVPR} & CVPR 2017 & RGB + LiDAR& 62.68 \\
				ContFuse\cite{Liang2018ECCV} & ECCV 2018  & RGB + LiDAR&  73.25  \\
				AVOD-FPN \cite{ku2018joint} & IROS 2018 & RGB + LiDAR & 74.44  \\			
				F-PointNet \cite{qi2017frustum} & CVPR 2018  & RGB + LiDAR & 70.92 \\ 
				\hline 	
				VoxelNet \cite{zhou2018voxelnet} & CVPR 2018 & LiDAR only& 65.46\\
				SECOND \cite{yan2018second} & Sensors 2018 & LiDAR only& 76.48 \\
				PointRCNN \cite{shi2019pointrcnn}& CVPR 2019 & LiDAR only & 78.63\\
				Fast Point R-CNN \cite{Chen2019fastpointrcnn} & ICCV 2019 & LiDAR only & 79.00 \\
				STD \cite{std2019yang} & ICCV 2019 & LiDAR only & 79.80 \\
				\hline 
				PV-RCNN (Ours) & - & LiDAR only & \textbf{83.90} \\ 
				\hline
			\end{tabular}
		}
	\end{center}
\vspace{-0.2cm}
	\caption{Performance comparison on the moderate level car class of KITTI \emph{val} split with mAP calculated by 11 recall positions. }
	\label{tab:val}
	\vspace{-0.1cm}
\end{table}

\subsection{3D Detection on the KITTI Dataset}\label{sec:kitti}

To evaluate the proposed model's performance on the KITTI \emph{val} split, we train our model on the \emph{train} set and report the results on the \emph{val} set. To conduct evaluation on the \emph{test} set with the KITTI official test server, the model is trained with $80\%$ of all available \emph{train+val} data and the remaining $20\%$ data is used for validation.

\noindent
\textbf{Evaluation Metric.}~
All results are evaluated by the mean average precision with a rotated IoU threshold $0.7$ for $cars$ and $0.5$ for $cyclists$. The mean average precisions on the \emph{test} set are calculated with $40$ recall positions on the official KITTI test server \cite{kitti_leaderboard}. The results on the \emph{val} set in Table~\ref{tab:val} are calculated with $11$ recall positions to compare with the results by the previous works.

\noindent
\textbf{Comparison with state-of-the-art methods.}~
Table~\ref{tab:test} shows the performance of PV-RCNN on the KITTI \textit{test} set from the official online leaderboard as of Nov.~15th, 2019. For the most important 3D object detection benchmark of the car class, our method outperforms previous state-of-the-art methods with remarkable margins, \ie increasing the mAP by 1.58\%, 1.72\%, 1.73\% on easy, moderate and hard difficulty levels, respectively. 
For the bird-view detection of the car class, our method also achieves new state-of-the-art performance on the easy and moderate difficulty levels while dropping slightly on the hard difficulty level. For 3D detection and bird-view detection of cyclist, our methods outperforms previous LiDAR-only methods with large margins on the moderate and hard difficulty levels while achieving comparable performance on the easy difficulty level. Note that we train a single model for both the car and cyclist detection instead of separate models for each class as previous methods \cite{yan2018second,lang2018pointpillars,shi2019pointrcnn,std2019yang} do.

As of Nov.~15th, 2019, our method currently ranks $1^{st}$ on the car 3D detection leaderboard among all methods including both the RGB+LiDAR methods and LiDAR-only methods, and ranks $1^{st}$ on the cyclist 3D detection leaderboard among all published LiDAR-only methods. The significant improvements manifest the effectiveness of the PV-RCNN.

We also report the performance of the most important car class on the KITTI \textit{val} split with mAP from $R11$. Similarly, as shown in Table~\ref{tab:val}, our method outperforms previous state-of-the-art methods with large margins. The performance with $R40$ are also provided in Table~\ref{tab:valR40} for reference.

\begin{table}
	\small 
	\begin{center}
		\scalebox{0.99}[0.9]{
			\setlength\tabcolsep{4.2pt}
			\begin{tabular}{c|ccc|ccc}
				\hline 
				\multirow{2}{*}{\tabincell{c}{IoU \\ Thresh.}} & 
				\multicolumn{3}{c|}{3D mAP} & 
				\multicolumn{3}{c}{BEV mAP}\\
				& Easy & Moderate  & Hard & Easy & Moderate  & Hard \\
				\hline
				0.7 & 92.57 & 84.83 & 82.69 & 95.76 & 91.11 & 88.93 \\ 
				\hline
			\end{tabular}
		}
	\end{center}
	\vspace{-0.2cm}
	\caption{Performance on the KITTI \emph{val} split set with mAP calculated by 40 recall positions for car class.}
	\label{tab:valR40}
	\vspace{-0.1cm}
\end{table}

\begin{table}
	\small 
	\begin{center}
		\scalebox{0.9}{
			\setlength\tabcolsep{4.8pt}
			\begin{tabular}{c|ccc}
				\hline 
				Method & PointRCNN \cite{shi2019pointrcnn} & STD \cite{std2019yang} & PV-RCNN (Ours) \\
				\hline
				Recall (IoU=0.7) & 74.8 & 76.8 & 85.5 \\ 
				\hline
			\end{tabular}
		}
	\end{center}
	\vspace{-0.2cm}
	\caption{Recall of different proposal generation networks on the car class at moderate difficulty level of the KITTI \textit{val} split set.}
	\label{tab:val_recall}
	\vspace{-0.3cm}
\end{table}

\begin{table*}
	\small 
	\vspace{-1mm}
	\begin{center}
		\scalebox{0.82}{
			\setlength\tabcolsep{2pt}
			\begin{tabular}{c|c||cccc|cccc|cccc|cccc}
				\hline
				\multirow{2}{*}{Difficulty} & \multirow{2}{*}{Method} &
				\multicolumn{4}{c|}{3D mAP (IoU=0.7)} & \multicolumn{4}{c|}{3D mAPH (IoU=0.7)} & 
				\multicolumn{4}{c|}{BEV mAP (IoU=0.7)} & 
				\multicolumn{4}{c}{BEV mAPH (IoU=0.7)} \\
				& & Overall & 0-30m & 30-50m & 50m-Inf & 
				Overall & 0-30m & 30-50m & 50m-Inf  & 
				Overall & 0-30m & 30-50m & 50m-Inf & 
				Overall & 0-30m & 30-50m & 50m-Inf \\
				\hline 
				\multirow{3}{*}{LEVEL 1} 
				& PointPillar \cite{lang2018pointpillars} & 56.62 & 81.01 & 51.75 & 27.94 & - & - & - & - & 75.57 & 92.1 & 74.06 & 55.47 & - & - & - & -  \\ 
				& MVF \cite{zhou2019end} & 62.93 & 86.30 & 60.02 & 36.02 & - & - & - & - & 80.40 & 93.59 & 79.21 & 63.09 & - & - & - & -\\
				\cline{2-18}
				& PV-RCNN (Ours) & \textbf{70.30} & \textbf{91.92} & \textbf{69.21} & \textbf{42.17} & \textbf{69.69} & \textbf{91.34} & \textbf{68.53} & \textbf{41.31} & \textbf{82.96} & \textbf{97.35} & \textbf{82.99} & \textbf{64.97} & \textbf{82.06} & \textbf{96.71} & \textbf{82.01} & \textbf{63.15} \\ 
				& \cellcolor{LightCyan}\textit{Improvement} & \cellcolor{LightCyan}\textit{+7.37} & \cellcolor{LightCyan}\textit{+5.62} & \cellcolor{LightCyan}\textit{+9.19} & \cellcolor{LightCyan}\textit{+6.15} & \cellcolor{LightCyan}- & \cellcolor{LightCyan}- & \cellcolor{LightCyan}- & \cellcolor{LightCyan}- & \cellcolor{LightCyan}\textit{+2.56} & \cellcolor{LightCyan}\textit{+3.76} & \cellcolor{LightCyan}\textit{+3.78} & \cellcolor{LightCyan}\textit{+1.88} & \cellcolor{LightCyan}- & \cellcolor{LightCyan}- & \cellcolor{LightCyan}- & \cellcolor{LightCyan}- \\
				\hline 
				LEVEL 2 & PV-RCNN (Ours) & 65.36 & 91.58 & 65.13 & 36.46 & 64.79 & 91.00 & 64.49 & 35.70 & 77.45 & 94.64 & 80.39 & 55.39 &76.60 & 94.03 & 79.40 & 53.82\\
				\hline
			\end{tabular}
		}
	\end{center}
	\vspace{-0.3cm}
	\caption{Performance comparison on the Waymo Open Dataset (version 1.0 released in August, 2019) with 202 validation sequences for the vehicle detection. Note that the results of PointPillar~\cite{lang2018pointpillars} on the Waymo Open Dataset are reproduced by \cite{zhou2019end}.}
	\label{tab:waymo3d}
\end{table*}

\begin{table*}
	\begin{center}
		\scalebox{0.725}{
			\begin{tabular}{c|c||cc|cc|cc|cc|cc|ccl}
				\hline
				\multirow{2}{*}{Method} & 
				\multirow{2}{*}{Reference} & 			
				\multicolumn{2}{c|}{~~ Vehicle (LEVEL 1)  ~~} & \multicolumn{2}{c|}{~Vehicle (LEVEL 2)~} & 
				\multicolumn{2}{c|}{~~Ped. (LEVEL 1) ~~} & \multicolumn{2}{c|}{~Ped. (LEVEL 2)~} & 
				\multicolumn{2}{c|}{~Cyc. (LEVEL 1)~} & \multicolumn{2}{c}{Cyc. (LEVEL 2)}\\
				&&mAP & mAPH & mAP & \underline{mAPH} & mAP & mAPH & mAP & \underline{mAPH} & mAP & mAPH & mAP & \underline{mAPH}\\
				\hline 
				*StarNet \cite{ngiam2019starnet} & NeurIPSw 2019 & 53.70 & - & - & - & 66.80 & - & - & - & - & - & - & - \\ 
				*PointPillar \cite{lang2018pointpillars} & CVPR 2019 & 56.62 & - & - & - & 59.25 & - & - & - & - & - & - & -\\ 
				*MVF \cite{zhou2019end} & CoRL 2019 & 62.93 & - & - & - & 65.33 & - & - & - & - & - & - & - \\ 
				$^\dagger$SECOND~\cite{yan2018second} & Sensors 2018 & 72.27 & 71.69 & 63.85 & 63.33 & 68.70 & 58.18 & 60.72 & 51.31 & 	60.62 & 59.28 & 58.34 & 57.05 \\ 
				\hline
				PV-RCNN (Ours) & - & \textbf{77.51} & \textbf{76.89} & \textbf{68.98} & \textbf{68.41}	& \textbf{75.01} & \textbf{65.65} & \textbf{66.04} & \textbf{57.61} & \textbf{67.81} & \textbf{66.35}	 & \textbf{65.39} & \textbf{63.98}\\ 
				\hline
			\end{tabular}
		}
	\end{center}
	\caption{Performance comparison on the Waymo Open Dataset (version 1.2 released in March 2020) with 202 validation sequences for three categories. $\dagger$: re-implemented by ourselves with their open source code. $*$: performance on the version 1.0 of Waymo Open Dataset.}
	\label{tab:waymo_val}
	\vspace{-3mm}
\end{table*} 

\subsection{3D Detection on the Waymo Open Dataset}\label{sec:waymo}

To further validate the effectiveness of our proposed PV-RCNN, we evaluate the performance of PV-RCNN on the newly released large-scale Waymo Open Dataset. 

\noindent
\textbf{Evaluation Metric.}~
We adopt the official released evaluation tools for evaluating our method, where the mean average precision (mAP) and the mean average precision weighted by heading (mAPH) are used for evaluation. The rotated IoU threshold is set as $0.7$ for vehicle detection and $0.5$ for pedestrian~/~cyclist.
The test data are split in two ways. The first way is based on objects' different distances to the sensor: $0-30m$, $30-50m$ and $>50m$. The second way is to split the data into  
two difficulty levels, where the LEVEL\_1 denotes the ground-truth objects with at least 5 inside points while the LEVEL\_2 denotes the ground-truth objects with at least 1 inside points or the ground-truth objects manually marked as LEVEL\_2. 

\noindent
\textbf{Comparison with state-of-the-art methods.}~
Table~\ref{tab:waymo3d} shows that our method outperforms previous state-of-the-art \cite{zhou2019end} significantly with a $7.37\%$ mAP gain for the 3D object detection and a $2.56\%$ mAP gain for the bird-view object detection. 
The results show that our method achieves remarkably better mAP on all distance ranges of interest, where the maximum gain is $9.19\%$ for the 3D detection in the range of $30-50m$, which validates that our proposed multi-level point-voxel integration strategy is able to effectively capture more accurate contextual information for improving the 3D detection performance. 
As shown in Table~\ref{tab:waymo3d}, our method also achieves superior performance in terms of mAPH, 
which demonstrates that our model predicted accurate heading direction for the vehicles. 
The results on the LEVEL\_2 difficult level are also reported in Table~\ref{tab:waymo3d} for reference, and we could see that our method performs well even for the objects with fewer than 5 inside points. 
The experimental results on the large-scale Waymo Open dataset further 
validate the generalization ability of our proposed framework on various datasets.

\noindent
\textbf{Better performance for multi-class detection with more proposals.}~
To evaluate the performance of our method for multi-class detection, we further conduct experiments on the latest Waymo Open Dataset (version 1.2 released in March 2020).  
Here the number of proposals is increased from 100 to 500 since we only train a single model for detecting all three categories (\eg, vehicle, pedestrian and cyclist). 
As shown in Table~\ref{tab:waymo_val}, our method significantly surpasses previous methods on all difficulty levels of these three categories. 
We hope it could set up a strong baseline on the Waymo Open Dataset for future works.

\begin{table}
	\small 
	\begin{center}
		\scalebox{0.85}[0.8]{
			\setlength\tabcolsep{2pt}
			\begin{tabular}{c|ccc||ccc}
				\hline
				Method & 
				\tabincell{c}{RPN with 3D \\Voxel CNN} & 
				\tabincell{c}{Keypoints \\ Encoding} &
				\tabincell{c}{RoI-grid \\ Pooling} 
				& Easy & Mod. & Hard \\
				\hline
				RPN Baseline & \checkmark & & &  90.46 & 80.87 & 77.30  \\
				Pool from Encoder& \checkmark & & \checkmark  & 91.88 & 82.86  & 80.52 \\
				PV-RCNN & \checkmark & \checkmark & \checkmark  &  \textbf{92.57} & \textbf{84.83} & \textbf{82.69} \\
				\hline
			\end{tabular}
		}
	\end{center}
	\vspace{-0.2cm}
	\caption{Effects of voxel-to-keypoint scene encoding strategy and RoI-grid pooling refinement.}
	\label{tab:exp_baseline}
	\vspace{-0.1cm}
\end{table}

\begin{table}
	\small 
	\begin{center}
		\scalebox{0.83}[0.75]{
			\begin{tabular}{cccc|cc||c}
				\hline
				$f_i^{(pv_1)}$ & $f_i^{(pv_2)}$ & $f_i^{(pv_3)}$ & $f_i^{(pv_4)}$ & $f_i^{(bev)}$ & 
				$f_i^{(raw)}$ 
				& \tabincell{c}{Moderate mAP}  \\
				\hline
				&&&&&\checkmark & 81.98 \\
				&&&&\checkmark& & 83.32\\
				&&&\checkmark&&& 83.17  \\
				&&\checkmark&\checkmark&&& 84.54  \\
				&&\checkmark&\checkmark&\checkmark&& 84.69  \\
				&&\checkmark&\checkmark&\checkmark&\checkmark& 84.72 \\
				&\checkmark&\checkmark&\checkmark&\checkmark&\checkmark& 84.75 \\
				\checkmark&\checkmark&\checkmark&\checkmark&\checkmark&\checkmark& \textbf{84.83} \\
				\hline
			\end{tabular}
		}
	\end{center}
	\vspace{-0.2cm}
	\caption{Effects of different feature components for VSA module.}
	\label{tab:exp_sa_features}
	\vspace{-0.4cm}
\end{table}

\subsection{Ablation Studies}\label{sec:ab}
In this section, we conduct extensive ablation experiments to analyze individual components of our proposed method. All models are trained on the \textit{train} split and evaluated on the \textit{val} split for the car class of KITTI dataset \cite{Geiger2012CVPR}.

\noindent
\textbf{Effects of voxel-to-keypoint scene encoding.}~
We validate the effectiveness of voxel-to-keypoint scene encoding strategy by comparing with the native solution that directly aggregating multi-scale feature volumes of encoder to the RoI-grid points as mentioned in Sec.~\ref{sec:rpn}. As shown in the $2^{nd}$ and $3^{rd}$ rows of Table~\ref{tab:exp_baseline}, the voxel-to-keypoint scene encoding strategy contributes significantly to the performance in all three difficulty levels. This benefits from that the keypoints enlarge the receptive fields by bridging the 3D voxel CNN and RoI-grid points, and the segmentation supervision of keypoints also enables a better multi-scale feature learning from the 3D voxel CNN. Besides, a small set of keypoints as the intermediate feature representation also decreases the GPU memory usage when compared with the directly pooling strategy.

\noindent
\textbf{Effects of different features for VSA module.}~
In Table~\ref{tab:exp_sa_features}, we investigate the importance of each feature component of keypoints in Eq.~\eqref{eq:keypointfeature0} and Eq.~\eqref{eq:keypointfeatures1}. The $1^{st}$ row shows that the performance drops a lot if we only aggregate features from $f_i^{(raw)}$, since the shallow semantic information is not enough for the proposal refinement. The high level semantic information from $f_i^{(pv_3)}$, $f_i^{(pv_4)}$ and $f_i^{(bev)}$ improves the performance significantly as shown in $2^{nd}$ to $5^{th}$ rows. As shown in last four rows, the additions of relative shallow semantic features $f_i^{(pv_1)}$, $f_i^{(pv_2)}$, $f_i^{(raw)}$ 
further improves the performance slightly and the best performance is achieved with all the feature components as the keypoint features.

\noindent
\textbf{Effects of PKW module.}~
We propose the predicted keypoint weighting (PKW) module in Sec.~\ref{voxel_to_point} to re-weight the point-wise features of keypoint with extra keypoint segmentation supervision. 
Table~\ref{tab:ab_study} ($1^{st}$ and $4^{th}$ rows) shows that removing the PKW module drops performance a lot, which demonstrates that the PKW module enables better multi-scale feature aggregation by focusing more on the foreground keypoints, since they are more important for the succeeding proposal refinement network.

\noindent
\textbf{Effects of RoI-grid pooling module.}~
We investigate the effects of RoI-grid pooling module by replacing it with the RoI-aware pooling \cite{shi2019part} and keeping the other modules consistent. Table~\ref{tab:ab_study} shows that the performance drops significantly when replacing RoI-grid pooling module, which validates that our proposed set abstraction based RoI-grid pooling could learn much richer contextual information,  and the pooled features also encode more discriminative RoI features by pooling more effective features with large search radii for each grid point. $1^{st}$ and $2^{nd}$ rows of Table~\ref{tab:exp_baseline} also shows that comparing with the 3D voxel RPN, the performance increases a lot after the proposal is refined by the features aggregated from the RoI-grid pooling module.

\begin{table}
	\small 
	\begin{center}
		\scalebox{0.9}{
			\setlength\tabcolsep{2pt}
			\begin{tabular}{ccc|ccc}
				\hline
				PKW & \tabincell{c}{RoI\\Pooling} & \tabincell{c}{Confidence\\Prediction}
				& Easy & Moderate & Hard \\
				\hline
				\cellcolor{LightRed}\xmark&RoI-grid Pooling&IoU-guided scoring &  92.09 & 82.95  & 81.93 \\
				\checkmark&\cellcolor{LightRed}RoI-aware Pooling&IoU-guided scoring & 92.54 & 82.97  & 80.30 \\
				\checkmark&RoI-grid Pooling&\cellcolor{LightRed}Classification &  91.71 & 82.50  & 81.41 \\
				\rowcolor{LightCyan}\checkmark&RoI-grid Pooling &IoU-guided Scoring &  \textbf{92.57} & \textbf{84.83} & \textbf{82.69} \\
				\hline
			\end{tabular}
		}
	\end{center}
	\vspace{-0.2cm}
	\caption{Effects of predicted keypoint weighting module, RoI-grid pooling module and IoU-guided confidence prediction.}
	\label{tab:ab_study}
	\vspace{-0.5cm}
\end{table}

\section{Conclusion}
We have presented the PV-RCNN framework, a novel method for accurate 3D object detection from point clouds. Our method integrates both the multi-scale 3D voxel CNN features and the PointNet-based features to a small set of keypoints by the new proposed voxel set abstraction layer, and the learned discriminative features of keypoints are then aggregated to the RoI-grid points with multiple receptive fields to capture much richer context information for the fine-grained proposal refinement. Experimental results on the KITTI dataset and the Waymo Open dataset demonstrate that our proposed voxel-to-keypoint scene encoding and keypoint-to-grid RoI feature abstraction strategy significantly improve the 3D object detection performance compared with previous state-of-the-art methods. 

{\small
\bibliographystyle{ieee_fullname}
\bibliography{egbib}

\begin{thebibliography}{10}\itemsep=-1pt

\bibitem{Chen2017CVPR}
Xiaozhi Chen, Huimin Ma, Ji Wan, Bo Li, and Tian Xia.
\newblock Multi-view 3d object detection network for autonomous driving.
\newblock In {\em The IEEE Conference on Computer Vision and Pattern
  Recognition (CVPR)}, July 2017.

\bibitem{Chen2019fastpointrcnn}
Yilun Chen, Shu Liu, Xiaoyong Shen, and Jiaya Jia.
\newblock Fast point r-cnn.
\newblock In {\em Proceedings of the IEEE international conference on computer
  vision (ICCV)}, 2019.

\bibitem{choy20194d}
Christopher Choy, JunYoung Gwak, and Silvio Savarese.
\newblock 4d spatio-temporal convnets: Minkowski convolutional neural networks.
\newblock In {\em Proceedings of the IEEE Conference on Computer Vision and
  Pattern Recognition}, pages 3075--3084, 2019.

\bibitem{Geiger2012CVPR}
Andreas Geiger, Philip Lenz, and Raquel Urtasun.
\newblock Are we ready for autonomous driving? the kitti vision benchmark
  suite.
\newblock In {\em Conference on Computer Vision and Pattern Recognition
  (CVPR)}, 2012.

\bibitem{3DSemanticSegmentationWithSubmanifoldSparseConvNet}
Benjamin Graham, Martin Engelcke, and Laurens van~der Maaten.
\newblock 3d semantic segmentation with submanifold sparse convolutional
  networks.
\newblock {\em CVPR}, 2018.

\bibitem{SubmanifoldSparseConvNet}
Benjamin Graham and Laurens van~der Maaten.
\newblock Submanifold sparse convolutional networks.
\newblock {\em CoRR}, abs/1706.01307, 2017.

\bibitem{huang2018recurrent}
Qiangui Huang, Weiyue Wang, and Ulrich Neumann.
\newblock Recurrent slice networks for 3d segmentation of point clouds.
\newblock In {\em Proceedings of the IEEE Conference on Computer Vision and
  Pattern Recognition}, pages 2626--2635, 2018.

\bibitem{jaritz2019multi}
Maximilian Jaritz, Jiayuan Gu, and Hao Su.
\newblock Multi-view pointnet for 3d scene understanding.
\newblock In {\em Proceedings of the IEEE International Conference on Computer
  Vision Workshops}, pages 0--0, 2019.

\bibitem{jiang2018acquisition}
Borui Jiang, Ruixuan Luo, Jiayuan Mao, Tete Xiao, and Yuning Jiang.
\newblock Acquisition of localization confidence for accurate object detection.
\newblock In {\em Proceedings of the European Conference on Computer Vision
  (ECCV)}, pages 784--799, 2018.

\bibitem{kitti_leaderboard}
{KITTI leader board of 3D object detection benchmark}.
\newblock
  \url{http://www.cvlibs.net/datasets/kitti/eval_object.php?obj_benchmark=3d},
  Accessed on 2019-11-15.

\bibitem{ku2018joint}
Jason Ku, Melissa Mozifian, Jungwook Lee, Ali Harakeh, and Steven Waslander.
\newblock Joint 3d proposal generation and object detection from view
  aggregation.
\newblock {\em IROS}, 2018.

\bibitem{lang2018pointpillars}
Alex~H Lang, Sourabh Vora, Holger Caesar, Lubing Zhou, Jiong Yang, and Oscar
  Beijbom.
\newblock Pointpillars: Fast encoders for object detection from point clouds.
\newblock {\em CVPR}, 2019.

\bibitem{lehner2019patch}
Johannes Lehner, Andreas Mitterecker, Thomas Adler, Markus Hofmarcher, Bernhard
  Nessler, and Sepp Hochreiter.
\newblock Patch refinement - localized 3d object detection.
\newblock {\em CoRR}, abs/1910.04093, 2019.

\bibitem{li2019gs3d}
Buyu Li, Wanli Ouyang, Lu Sheng, Xingyu Zeng, and Xiaogang Wang.
\newblock Gs3d: An efficient 3d object detection framework for autonomous
  driving.
\newblock In {\em Proceedings of the IEEE Conference on Computer Vision and
  Pattern Recognition}, pages 1019--1028, 2019.

\bibitem{li2018pointcnn}
Yangyan Li, Rui Bu, Mingchao Sun, Wei Wu, Xinhan Di, and Baoquan Chen.
\newblock Pointcnn: Convolution on x-transformed points.
\newblock In {\em Advances in Neural Information Processing Systems}, pages
  820--830, 2018.

\bibitem{Liang2019CVPR}
Ming Liang*, Bin Yang*, Yun Chen, Rui Hu, and Raquel Urtasun.
\newblock Multi-task multi-sensor fusion for 3d object detection.
\newblock In {\em CVPR}, 2019.

\bibitem{Liang2018ECCV}
Ming Liang, Bin Yang, Shenlong Wang, and Raquel Urtasun.
\newblock Deep continuous fusion for multi-sensor 3d object detection.
\newblock In {\em ECCV}, 2018.

\bibitem{lin2018focal}
Tsung-Yi Lin, Priyal Goyal, Ross Girshick, Kaiming He, and Piotr Doll{\'a}r.
\newblock Focal loss for dense object detection.
\newblock {\em IEEE transactions on pattern analysis and machine intelligence},
  2018.

\bibitem{liu2019point}
Zhijian Liu, Haotian Tang, Yujun Lin, and Song Han.
\newblock Point-voxel {CNN} for efficient 3d deep learning.
\newblock {\em CoRR}, abs/1907.03739, 2019.

\bibitem{ngiam2019starnet}
Jiquan Ngiam, Benjamin Caine, Wei Han, Brandon Yang, Yuning Chai, Pei Sun, Yin
  Zhou, Xi Yi, Ouais Alsharif, Patrick Nguyen, Zhifeng Chen, Jonathon Shlens,
  and Vijay Vasudevan.
\newblock Starnet: Targeted computation for object detection in point clouds.
\newblock {\em CoRR}, abs/1908.11069, 2019.

\bibitem{qi2019deep}
Charles~R. Qi, Or Litany, Kaiming He, and Leonidas~J. Guibas.
\newblock Deep hough voting for 3d object detection in point clouds.
\newblock In {\em The IEEE International Conference on Computer Vision (ICCV)},
  October 2019.

\bibitem{qi2017frustum}
Charles~R. Qi, Wei Liu, Chenxia Wu, Hao Su, and Leonidas~J. Guibas.
\newblock Frustum pointnets for 3d object detection from rgb-d data.
\newblock In {\em The IEEE Conference on Computer Vision and Pattern
  Recognition (CVPR)}, June 2018.

\bibitem{qi2017pointnet}
Charles~R Qi, Hao Su, Kaichun Mo, and Leonidas~J Guibas.
\newblock Pointnet: Deep learning on point sets for 3d classification and
  segmentation.
\newblock In {\em Proceedings of the IEEE Conference on Computer Vision and
  Pattern Recognition}, pages 652--660, 2017.

\bibitem{qi2017pointnet++}
Charles~Ruizhongtai Qi, Li Yi, Hao Su, and Leonidas~J Guibas.
\newblock Pointnet++: Deep hierarchical feature learning on point sets in a
  metric space.
\newblock In {\em Advances in Neural Information Processing Systems}, pages
  5099--5108, 2017.

\bibitem{shi2019pointrcnn}
Shaoshuai Shi, Xiaogang Wang, and Hongsheng Li.
\newblock Pointrcnn: 3d object proposal generation and detection from point
  cloud.
\newblock In {\em Proceedings of the IEEE Conference on Computer Vision and
  Pattern Recognition}, pages 770--779, 2019.

\bibitem{shi2019part}
Shaoshuai Shi, Zhe Wang, Jianping Shi, Xiaogang Wang, and Hongsheng Li.
\newblock From points to parts: 3d object detection from point cloud with
  part-aware and part-aggregation network.
\newblock {\em IEEE Transactions on Pattern Analysis and Machine Intelligence},
  2020.

\bibitem{song2016deep}
Shuran Song and Jianxiong Xiao.
\newblock Deep sliding shapes for amodal 3d object detection in rgb-d images.
\newblock In {\em Proceedings of the IEEE Conference on Computer Vision and
  Pattern Recognition}, pages 808--816, 2016.

\bibitem{su2018splatnet}
Hang Su, Varun Jampani, Deqing Sun, Subhransu Maji, Evangelos Kalogerakis,
  Ming-Hsuan Yang, and Jan Kautz.
\newblock Splatnet: Sparse lattice networks for point cloud processing.
\newblock In {\em Proceedings of the IEEE Conference on Computer Vision and
  Pattern Recognition}, pages 2530--2539, 2018.

\bibitem{thomas2019kpconv}
Hugues Thomas, Charles~R. Qi, Jean-Emmanuel Deschaud, Beatriz Marcotegui,
  Francois Goulette, and Leonidas~J. Guibas.
\newblock Kpconv: Flexible and deformable convolution for point clouds.
\newblock In {\em The IEEE International Conference on Computer Vision (ICCV)},
  October 2019.

\bibitem{wang2019voxelFPN}
Bei Wang, Jianping An, and Jiayan Cao.
\newblock Voxel-fpn: multi-scale voxel feature aggregation in 3d object
  detection from point clouds.
\newblock {\em CoRR}, abs/1907.05286, 2019.

\bibitem{wang2019dynamic}
Yue Wang, Yongbin Sun, Ziwei Liu, Sanjay~E Sarma, Michael~M Bronstein, and
  Justin~M Solomon.
\newblock Dynamic graph cnn for learning on point clouds.
\newblock {\em ACM Transactions on Graphics (TOG)}, 38(5):146, 2019.

\bibitem{wang2019frustum}
Zhixin Wang and Kui Jia.
\newblock Frustum convnet: Sliding frustums to aggregate local point-wise
  features for amodal 3d object detection.
\newblock In {\em IROS}. IEEE, 2019.

\bibitem{wu2019pointconv}
Wenxuan Wu, Zhongang Qi, and Li Fuxin.
\newblock Pointconv: Deep convolutional networks on 3d point clouds.
\newblock In {\em Proceedings of the IEEE Conference on Computer Vision and
  Pattern Recognition}, pages 9621--9630, 2019.

\bibitem{yan2018second}
Yan Yan, Yuxing Mao, and Bo Li.
\newblock Second: Sparsely embedded convolutional detection.
\newblock {\em Sensors}, 18(10):3337, 2018.

\bibitem{Yang2018CoRL}
Bin Yang, Ming Liang, and Raquel Urtasun.
\newblock Hdnet: Exploiting hd maps for 3d object detection.
\newblock In {\em 2nd Conference on Robot Learning (CoRL)}, 2018.

\bibitem{yang2018pixor}
Bin Yang, Wenjie Luo, and Raquel Urtasun.
\newblock Pixor: Real-time 3d object detection from point clouds.
\newblock In {\em Proceedings of the IEEE Conference on Computer Vision and
  Pattern Recognition}, pages 7652--7660, 2018.

\bibitem{std2019yang}
Zetong Yang, Yanan Sun, Shu Liu, Xiaoyong Shen, and Jiaya Jia.
\newblock {STD:} sparse-to-dense 3d object detector for point cloud.
\newblock {\em ICCV}, 2019.

\bibitem{zhao2019pointweb}
Hengshuang Zhao, Li Jiang, Chi-Wing Fu, and Jiaya Jia.
\newblock Pointweb: Enhancing local neighborhood features for point cloud
  processing.
\newblock In {\em Proceedings of the IEEE Conference on Computer Vision and
  Pattern Recognition}, pages 5565--5573, 2019.

\bibitem{zhou2019}
Dingfu Zhou, Jin Fang, Xibin Song, Chenye Guan, Junbo Yin, Yuchao Dai, and
  Ruigang Yang.
\newblock Iou loss for 2d/3d object detection.
\newblock In {\em International Conference on 3D Vision (3DV)}. IEEE, 2019.

\bibitem{zhou2019end}
Yin Zhou, Pei Sun, Yu Zhang, Dragomir Anguelov, Jiyang Gao, Tom Ouyang, James
  Guo, Jiquan Ngiam, and Vijay Vasudevan.
\newblock End-to-end multi-view fusion for 3d object detection in lidar point
  clouds.
\newblock {\em CoRR}, abs/1910.06528, 2019.

\bibitem{zhou2018voxelnet}
Yin Zhou and Oncel Tuzel.
\newblock Voxelnet: End-to-end learning for point cloud based 3d object
  detection.
\newblock In {\em Proceedings of the IEEE Conference on Computer Vision and
  Pattern Recognition}, pages 4490--4499, 2018.

\bibitem{zhu2019class}
Benjin Zhu, Zhengkai Jiang, Xiangxin Zhou, Zeming Li, and Gang Yu.
\newblock Class-balanced grouping and sampling for point cloud 3d object
  detection.
\newblock {\em CoRR}, abs/1908.09492, 2019.

\end{thebibliography}
}

\end{document}